\documentclass[sigconf, screen]{acmart}

\AtBeginDocument{%
  \providecommand\BibTeX{{%
    \normalfont B\kern-0.5em{\scshape i\kern-0.25em b}\kern-0.8em\TeX}}}

\copyrightyear{2020} 
\acmYear{2020} 
\setcopyright{acmlicensed}\acmConference[VINCI 2020]{The 13th International Symposium on Visual Information Communication and Interaction}{December 8--10, 2020}{Eindhoven, Netherlands}
\acmBooktitle{The 13th International Symposium on Visual Information Communication and Interaction (VINCI 2020), December 8--10, 2020, Eindhoven, Netherlands}
\acmPrice{15.00}
\acmDOI{10.1145/3430036.3430050}
\acmISBN{978-1-4503-8750-7/20/12}

\usepackage{amsmath}
\usepackage{hyperref}
\usepackage{csquotes}
\usepackage{listings}
\usepackage{enumitem}
\usepackage{subcaption}
\usepackage[capitalize]{cleveref}
\newcommand{\eqns}[1]{\begin{flalign} #1 \end{flalign}}
\newcommand{\eqnsnn}[1]{\begin{flalign*} #1 \end{flalign*}}

\newcommand{\T}{^\top}

\mathcode`\*="8000
{\catcode`\*\active\gdef*{\cdot}}
\crefname{lstlisting}{listing}{listings}
\Crefname{lstlisting}{Listing}{Listings}
\newcommand{\showchange}[1]{{#1}}
\graphicspath{{figures/}{pictures/}{images/}{./}} % where to search for the images

\begin{document}

\title[Visualization of Nonlinear Programming for Robot Motion Planning]{Visualization of Nonlinear Programming\\for Robot Motion Planning}

\author{David H{\"a}gele}
\affiliation{%
	\institution{University of Stuttgart}
	\streetaddress{Allmandring 19}
	\city{Stuttgart}
	\country{Germany}
	\postcode{70569}}
\email{david.haegele@visus.uni-stuttgart.de}

\author{Moataz Abdelaal}
\affiliation{%
	\institution{University of Stuttgart}
	\streetaddress{Allmandring 19}
	\city{Stuttgart}
	\country{Germany}
	\postcode{70569}}
\email{moataz.abdelaal@visus.uni-stuttgart.de}

\author{Ozgur S. Oguz}
\affiliation{%
	\institution{University of Stuttgart}
	\streetaddress{Universit{\"a}tsstraße 38}
	\city{Stuttgart}
	\country{Germany}
	\postcode{70569}}
\email{ozgur.oguz@ipvs.uni-stuttgart.de}

\author{Marc Toussaint}
\affiliation{%
	\institution{Technische Universit{\"a}t Berlin}
	\city{Berlin}
	\country{Germany}}
\email{toussaint@tu-berlin.de}

\author{Daniel Weiskopf}
\affiliation{%
	\institution{University of Stuttgart}
	\streetaddress{Allmandring 19}
	\city{Stuttgart}
	\country{Germany}
	\postcode{70569}}
\email{daniel.weiskopf@visus.uni-stuttgart.de}

\renewcommand{\shortauthors}{H{\"a}gele et al.}

\begin{abstract}
Nonlinear programming targets nonlinear optimization with constraints, which is a generic yet complex methodology involving humans for problem modeling and algorithms for problem solving. 
We address the particularly hard challenge of supporting domain experts in handling, understanding, and trouble-shooting high-dimensional optimization with a large number of constraints. 
Leveraging visual analytics, users are supported in exploring the computation process of nonlinear constraint optimization. 
Our system was designed for robot motion planning problems and developed in tight collaboration with domain experts in nonlinear programming and robotics. 
We report on the experiences from this design study, illustrate the usefulness for relevant example cases, and discuss the extension to visual analytics for nonlinear programming in general.

\end{abstract}




\begin{teaserfigure}
  \includegraphics[width=\textwidth]{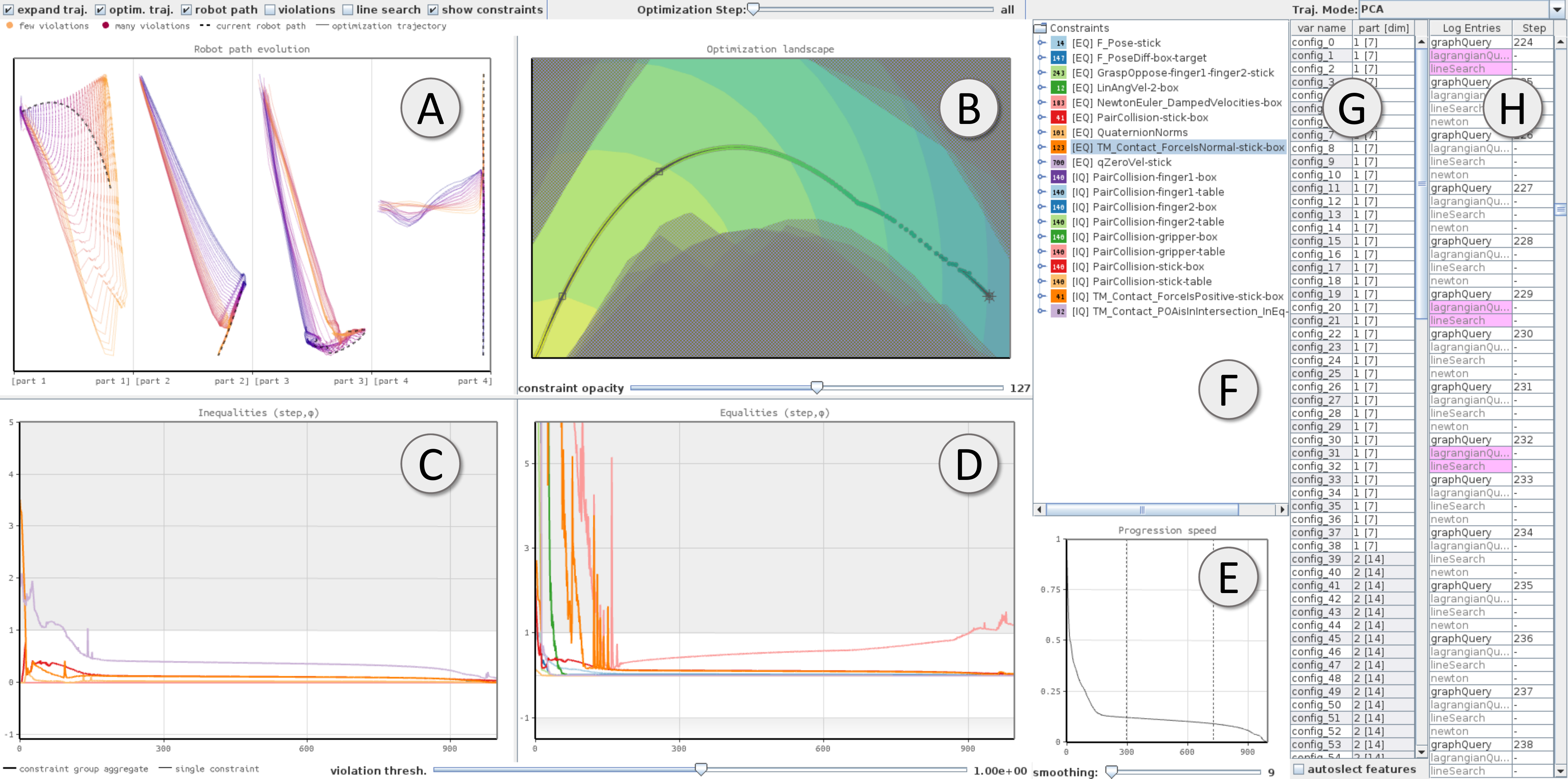}
  \caption{User interface of our visual analytics system for nonlinear programming processes. Using multiple coordinated views, a user can assess the evolution of a problem's solution, constraint values, and progression speed throughout an optimization.}
  \label{fig:teaser}
\end{teaserfigure}

\maketitle

\section{Introduction}
Nonlinear constraint optimization, also known as nonlinear programming (NLP), deals with finding optima within constrained sets of variables.
Only little work has been published concerning visualization in this field, which is surprising since a substantial amount of research has been conducted in the closely related fields of discrete and also unconstrained optimization.
NLP comprises two separate stages: the modeling stage and the solving stage.
In the modeling stage, the problem needs to be expressed in terms of an objective function and associated equality or inequality constraint functions that may be nonlinear.
A classical example from economics is to optimize spending for greatest profit (objective) while staying within budget (constraint).
In the solving stage, an iterative algorithm moves through the (likely high-dimensional) space of the modeled problem to find the optimal location with respect to the objective while making sure to satisfy all of the imposed constraints.

Due to high dimensionality, large number of constraints, and nonlinearity, which gives rise to local optima and disconnected feasible regions, it is challenging to grasp such optimization problems.
Even more challenging is the comprehension of unexpected or unsatisfactory behavior of a solver when applied to a problem.

To help NLP experts in understanding their problems better as well as the corresponding behavior of the applied solvers, we want to leverage visualization of the internal steps of an optimization process.
We propose a visual analytics approach for a post-mortem analysis of optimization runs of robot motion planning problems that we developed in tight collaboration with domain experts.
We focus on visualizing the high-dimensional optimization landscape as seen by the optimizer, to be able to reason about its behavior, and on representing the evolution of the solution throughout the optimization in order to be able to interpret high-dimensional loci as intermediate solutions to the motion planning problem. 
Our contributions are: 1) a visual analytics system for the analysis of constrained optimizations for robot motion planning, 2) a report of our design study process and lessons learned.

\section{Related Work}
This section is divided into two parts. 
First, we review related work in the field of visualization of optimization that includes NLP, linear programming, constraint programming, multi-objective optimization, as well as unconstrained optimization.  
Then, we discuss related work concerned with the visualization of temporal high-dimensional data.

\subsection{Visualization of Optimization}
\label{relatedworksub1}

Despite its strong potential~\cite{MESSAC2000, constraintprogrammingstudy}, the area of visualizing NLP remains unexplored to a large extent.
Androulakis and Vrahatis~\cite{ANDROULAKIS199641} proposed OPTAC, a tool for analyzing and visualizing the convergence behavior of unconstrained optimization algorithms.
Charalambos and Izquierdo~\cite{LPviz} visualize the geometric shapes of the feasible regions of linear programs. 
The method uses three-dimensional Cartesian coordinates and therefore, is limited to three-dimensional optimization problems. 
To display high-dimensional planes, Chatterjee et al.~\cite{LPparallelcoords} use parallel coordinates plots instead. 
Since the geometry in linear programs is simple and solvers are fundamentally different, these approaches are not applicable for NLP.

The area of constraint programming, in contrast to NLP, received a lot of attention from visualization practitioners. 
Most of the work done in this area focuses on visualizing the search tree resulting from constraint programs~\cite{carro2000apttool,Pu2000,cpviz,shishmarev2016visual,constraintprogrammingstudy}. 
However, these techniques cannot be applied to NLP, due to the differences in modeling and solvers~\cite{Heipcke1999}.
While solving constraint programs involves tree traversal and dynamic programming, NLPs consist of differentiable implicit surfaces, and their solvers are based on algebraic methods such as gradient descent. 

Instead of focusing on visualizing the search tree resulting from constraint programming, others attempt to visualize the evolution of variables, constraints, and the interaction between them using matrix views~\cite{carro2000VIFIDtool,Ghoniem2005Peeking,ghoniem2004visexp}. 
While matrices provide good representations to explore the relationship between constraints and variables, they have scalability issues, making them only suitable for exploring optimization problems with a small number of variables and constraints. 
Despite the differences, we find our work shares the same goal, i.e., exploring the evolution of variables and constraints.

The visualization of multi-objective optimization is yet another neighboring field to NLP. 
Most of the work in this field is concerned with visualizing the solution set in the objective space. 
Therefore, different visualization methods for high-dimensional data are used. 
Other methods apply dimensionality reduction to map the high-dimensional objective space into a two-dimensional space. 
Tu\v{s}ar and Filipi\v{c}~\cite{tuvsar2014visualization} provide a comprehensive review of these methods. 
Although multi-objective optimization and NLP address two different problems, they both share the notion of high dimensionality.

The visualization of high-dimensional unconstrained problem optimization such as in neural networks is discussed by Goodfellow et al.~\cite{characterizingnn}, who use a straight line from initialization to found optimum to sample and analyze the loss function.
This gave rise to the loss landscape visualization technique by Li~et~al.~\cite{losslandscape}. They use a 2D plane to sample the loss function and obtain a contour plot to analyze a subspace into which the optimizer's trajectory can be projected.
We extend this technique for NLP to show constraints within the landscape of the objective function.

\subsection{Visualization of Temporal High-Dimensional Data}
\label{relatedworksub2}

There are numerous methods for visualizing high-dimensional data~\cite{Liu2017}, scatterplot matrices~\cite{andrews1972plots}, glyphs~\cite{chernoff1973use}, parallel coordinates plots~\cite{inselberg1985plane}, and star coordinates~\cite{kandogan2000star}. 
These techniques, however, do not scale with a large number of dimensions. Therefore, other approaches project high-dimensional data into low-dimensional space using various dimensionality reduction techniques~\cite{nonato2018multidimensional}. 
Introducing the time dimension poses a visualization challenge, as the visual representation needs to convey not only the relation between the different dimensions but also their temporal context and evolution. 
Aigner et al.~\cite{aigner2011visualization} provide a comprehensive overview of such time-oriented visualization. 

A rather straightforward approach is to include time as an additional dimension in the traditional high-dimensional data representations. 
For example, Wong and Bergeron add an axis for the time dimension in parallel coordinates ~\cite{wong199430}. 
Similarly, TimeWheel~\cite{tominski2004axes} arranges the other axes in a circular layout around the time axis. 
These however, provide a poor representation of the temporal context information and are prone to become cluttered. 
Other approaches use a depth cue to encode the temporal information, resulting in 3D parallel coordinates~\cite{wegenkittl1997visualizing,gruendl2016time}, 3D star coordinates~\cite{noirhomme2002visualization}, or space-time cubes~\cite{spacetimecubeoperations,bach2015small}. 
These methods, however, have scalability concerns regarding the size of the data. 
Additionally, the use of interaction and animation is a necessity to avoid occlusion problems.

More recent approaches use dimensionality reduction to show temporal information. 
J\"ackle et al.~\cite{temporalmds} proposed temporal multidimensional scaling (TMDS) for visualizing multivariate time series data. 
Bach et al.~\cite{timecurves} and van den Elzen et al.~\cite{snapshotstopoints} showed the temporal progression of datasets by projecting the individual snapshots of datasets as points in 2D space. 
The points of subsequent snapshots are then connected by lines, while color is used to encode time.

In our work, we think of the optimization process as a sequence of intermediate solutions. 
Each solution is characterized by a set of high-dimensional decision variables. 
Thus, we can adopt the same techniques~\cite{timecurves,snapshotstopoints} to visualize the evolution of the optimization process.
In our work, in contrast, we deal with two different notions of time simultaneously: the time of a robot's motion and the time of the optimization process.

Torsney-Weir et al.~\cite{hyperslice} studied multi-dimensional shapes using their hypersliceplorer algorithm.
Using separate views for pairs of dimensions does not scale well with our problems, however.

\section{Background}
In this section, we summarize the domain background of robot motion planning and nonlinear programming.
We also present the requirements for a visualization system that we elicited from the domain experts. From here on we will use \emph{NLP} as an abbreviation for both, nonlinear program and nonlinear programming.

\subsection{Robot Motion Planning}
\begin{figure}[h]
	\centering
	\begin{subfigure}{0.32\linewidth}
		\includegraphics[width=\linewidth]{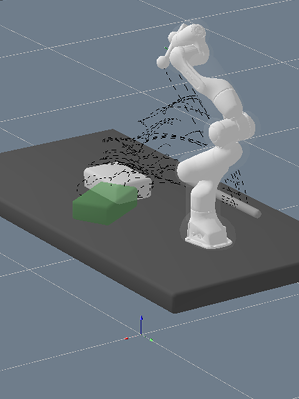}
	\end{subfigure}
	\begin{subfigure}{0.32\linewidth}
		\includegraphics[width=\linewidth]{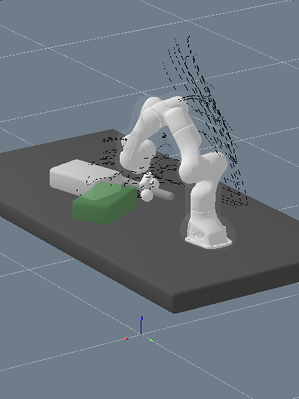}
	\end{subfigure}
	\begin{subfigure}{0.32\linewidth}
		\includegraphics[width=\linewidth]{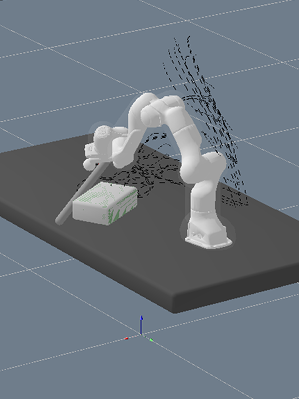}
	\end{subfigure}
	\caption{Animation frames from an optimized robot motion sequence for a box-pushing problem.
		The robot arm picks up a stick and then moves a box to a target location.
		The animation was created by our collaborators' motion planning framework.
	}
	\label{fig:plannedmotion}
\end{figure}

Motion planning is the problem of finding a collision-free path to move an object from an initial state to a desired goal state~\cite{latombe2012robot}.
It is a crucial problem in many fields, including robotics, computational biology (drug design, protein folding), virtual prototyping, manufacturing, and computer graphics. 
In this paper, we focus on robot motion planning as our application domain.
However, we expect that our visual analytics approach will carry over to other applications.
An example of such motion planning problems is illustrated in \cref{fig:plannedmotion}.

There are different approaches for finding a valid path connecting the source and target configurations.
LaValle~\cite{lavalle2006planning} provides an overview of the most important ones.
Grid- and sampling-based approaches~\cite{KONDO1991,SANDROS,Kavraki1994,Kavraki1996} discretize the configuration space and use graph search algorithms to find a valid path in the resulting topological space.

Another approach for finding a valid path is based on nonlinear optimization~\cite{toussaintmotionplanning}. 
This involves the formulation and solving of an NLP that describes the desired goal mathematically and imposes constraints to ensure a collision-free path and feasible motion.
In contrast to the aforementioned methods, nonlinear optimization does not require any discretization of the configuration space. 
Therefore, it is possible to find an optimal path if the problem is well-defined.

However, nonlinear optimization requires an initial guess of the solution and the method is prone to get stuck in local minima~\cite{chinneck2006practical}. 
To avoid that, it is common to use path finding algorithms as a first step to provide a reasonable initial guess. 
While this might be enough to find a valid solution, it does not provide insights into the internal mechanism of the optimization algorithms: 
How fast does the optimizer progress? Or which constraints hinder the algorithm the most? Such questions cannot be answered without a proper investigation of the inner workings of the optimization.

To this end, there is a critical need for visualization tools to debug and troubleshoot the optimization algorithms~\cite{MESSAC2000, constraintprogrammingstudy}. 
Having such tools, can help domain experts formulate hypotheses about the behavior of the optimizer, that could eventually lead to a better reformulation of the cost function and/or the constraints. 
In this context, we were approached by robotics researchers.
During the course of this design study, we have been collaborating with them for more than 9 months, integrating their expertise and knowledge about their domain-specific problems.
To better understand the domain problem we tried to characterize it by eliciting requirements for a visual analytics system and tasks to be achievable from the experts.
We were able to identify two high-level information needs:
\begin{enumerate}[label=Q{{\arabic*}}:, itemsep=0mm]
	\item When an optimization requires a lot of time to converge, what is taking the optimizer so long?
	\item When an optimization produces an infeasible solution, what prevents the optimizer from getting to a feasible location?
\end{enumerate}
To be able to analyze an optimization with regard to these questions, we formulated the following analysis tasks that the experts want to perform on optimizations of their robot motion planning framework:
\begin{enumerate}[label=T{{\arabic*}}:, itemsep=0mm]
	\item
	Identify and characterize different phases of the optimization process.
	
	\item
	Identify when the optimizer converges faster or more slowly to a local minimum during the optimization process.
	
	\item
	Characterize the evolution of the constraint function values during the optimization process.
	
	\item
	Identify the shape and boundaries of a constraint.%, to see where feasible regions are located.
	
	\item
	Identify the shape and boundaries of the feasible regions.
\end{enumerate}
Throughout the project we followed a design study process~\cite{sedlmair} regularly meeting with our collaboration partners to continuously refine our design and to update requirements.

\subsection{Nonlinear Programming}
Nonlinear programs consist of three parts that allow expressing an optimization problem with auxiliary conditions on the solution, known as the constraints. It typically reads like this:
\eqns{
	\text{minimize}~f(x) ~~ \text{subject to}~ h_i(x) = 0 ~,~ g_j(x) \leq 0
}
In this, $x$ is the set of decision variables that can be varied.
It makes sense, in our case, to think about $x$ as a large vector describing a robot's motion.
The objective function $f$ is to be minimized while all of the inequality constraints $g_j$ and equality constraints $h_i$ have to be satisfied at the solution $x^\ast$. 
In an NLP, at least one of these functions is nonlinear.
In high-dimensional space, the equalities define hypersurfaces, and inequalities define hypersurface-confined sets in which the optimal $x^\ast$ has to be located.

Including time into an optimization problem is often done by discretization into several time steps.
This also applies to our case of robot motion, where the motion path consists of joint configurations representing the robot's poses over time.

The objective and constraints are used to express the robot's task mathematically and to guarantee that the resulting motion is physically viable.
In such time-dependent optimization problems, some constraints will typically apply to individual time steps only to enforce valid states, and others will take several times steps into account to enforce valid state transitions or global properties.

This kind of modeling leads to high-dimensional problem spaces and large numbers of constraints.
We are dealing with $\approx 100$ time discrete joint configurations that range from 10 to 30 dimensions, amounting in an optimization space of 1000- or more dimensions, and more than 100 constraints.

\paragraph{Technical Background:}
We now discuss a class of iterative algorithms to solve NLPs.
We will not go into details but sketch the general mechanic that is common in the log-barrier, squared penalty, and also the augmented Lagrangian methods.
All of these want to find a solution that satisfies the Karush-Kuhn-Tucker conditions~\cite{kuhntuckernlp} (a set of conditions that hold at a feasible minimum of an NLP), and thus determine the dual variables.
This is done by constructing an unconstrained problem $L_{\lambda,\kappa}(x)$ from the NLP that can be minimized using standard techniques like gradient descent or Newtons's method.
\showchange{For example in the augmented Lagrangian method $L_{\lambda,\kappa}(x) = f(x)+\kappa\T h(x) + \lambda\T g(x) + ||h(x)||_2^2 + \sum_j [g_j(x)>0]g_j(x)^2$.}
The unconstrained problem is then minimized repeatedly while updating the dual parameters $\lambda,\kappa$, and thus changing the impact of constraints, in between.

A pseudo code implementation is shown in \cref{lst:augmentedlagrangian}. 
A line search mechanism (most inner loop decreasing step size) is leveraged to ensure sufficient decrease in each step to prevent overshooting valleys and enforces satisfaction of the Wolfe condition~\cite{wolfe}.

\begin{minipage}{0.95\linewidth}
	\lstset{keywords={,INPUT, OUTPUT, BEGIN, END, DO, NOT, UNTIL, WHILE, FUNCTION, }
	}
\begin{lstlisting}[
	caption={Pseudo Code for NLP Solver},
	label={lst:augmentedlagrangian},
	mathescape=true,
	numbers=left
	]
FUNCTION solveNLP
INPUT: objective $f$, equalities $h$, and inequalities $g$
OUTPUT: optimal location $x$
BEGIN
 define unconstrained problem $L_{\lambda,\kappa}$ from $f$, $h$, and $g$
 initialize $x$ randomly (or informed)
 DO
  $x$ = argmin of $L_{\lambda,\kappa}$ with initialization $x$
  update dual parameters $\lambda,\kappa$
 UNTIL convergence
END

FUNCTION argmin
INPUT: loss function $L$, initialization $x$
OUTPUT: optimal location $x$
BEGIN
 initialize stepsize $a=1$ (or previous value)
 WHILE $a$ reasonably large
  set descent direction $\delta$ = Newton step for $L$ at $x$
  WHILE $L(x + a\delta)$ NOT sufficiently smaller than $L(x)$
   decrease stepsize $a$
  END WHILE
  $x = x + a\delta$
  increase stepsize $a$
 END WHILE
END
\end{lstlisting}
	\vspace{2ex}
\end{minipage}

Besides the problem formulation, i.e., functions $f$, $g$, and $h$, this algorithm is the source of our data for visualization.
We define the optimization trajectory as the sequence $S$ of arguments $x' = x+a\delta$ that are tested during line search in the \texttt{argmin} procedure of \cref{lst:augmentedlagrangian} (line 20):
\eqns{\label{eq:optimizationtrajectory}
	S=\{x_\text{init}, x_1, x_2, .., x^\ast\}
}
Furthermore, we obtain a detailed log of the sequence in which different stages in the algorithm were executed:
updates to the dual variables (line~9), evaluations of the loss function and corresponding function values of objective and constraints (line~20), step size decrease during line search (line~21), and updates to $x$ (line~23).

\section{Method}
In this section, we will introduce our visual analytics system and discuss the special techniques used to visualize the optimization process, i.e., the loss landscape visualization and the robot path evolution visualization.

\subsection{Visual Analytics System}
Our visual analytics system combines and links different views
to facilitate the exploration of an optimization process.
Due to the iterative nature of the algorithm, the optimization process describes the evolution of essentially two things:
The evolution of objective and constraint function values, and the evolution of the solution described by the sequence of intermediate solutions $s_i \in S$ on the optimization trajectory.

To show the evolution of function values, which allows for task T3 to be accomplished, we provide line charts for equality and inequality constraints, plotting $h(s_i)$ and $g(s_i)$ with optimization step $i$ on the $x$-axis (views C and D in \cref{fig:teaser}).
We group constraints by names due to their large number, which allows us to prevent visual clutter.
For each group $H_k$, $G_k$, we plot the maximum value of all its constraints as aggregation.
\eqnsnn{
	\bar{h}_k(s_i) = \underset{h_j \in H_k}{\text{max}}(h_j(s_i))~;~~\bar{g}_k(s_i) = \underset{g_j \in G_k}{\text{max}}(|g_j(s_i)|)
}
Groups are listed in a separate view (F) and can be expanded if desired, to reveal the individual unaggregated constraints in the plot.

\begin{figure}
	\centering
	\begin{subfigure}{0.48\linewidth}
		\includegraphics[width=\linewidth]{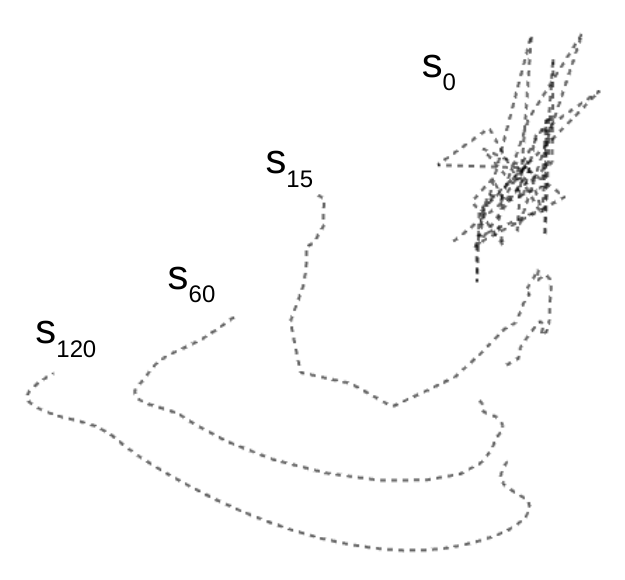}
		\caption{path evolution}
		\label{fig:pathevolution:paths}
	\end{subfigure}
	\begin{subfigure}{0.48\linewidth}
		\includegraphics[width=\linewidth]{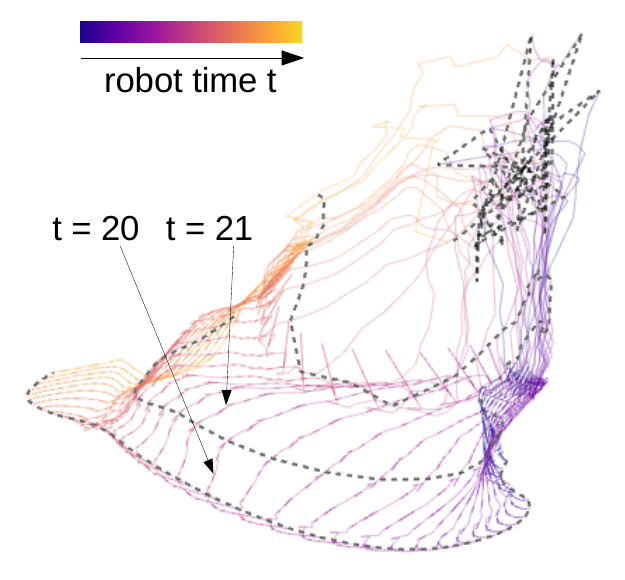}
		\caption{per configuration trajectories}
		\label{fig:pathevolution:trajectories}
	\end{subfigure}
	\caption{
		Figures show the motion path evolution over the optimization process using PCA for projection.
		(\subref{fig:pathevolution:paths}) It can be observed how the path evolves from a chaotic random shape (top) to a smooth curve (bottom).
		(\subref{fig:pathevolution:trajectories}) The trajectory from initialization to final solution for each configuration $c_t$ is shown and color-coded by time $t$.
		The underlying optimization corresponds to an NLP for pushing a box in a circular motion similar to the illustration in \cref{fig:plannedmotion}.
	}
	\label{fig:pathevolution}
\end{figure}

To show the evolution of the problem's solution we employ a time curves~\cite{timecurves} approach in which we reduce dimensionality of the joint configurations of $s = \{c_1 \hdots c_T\}$, that is the motion path (view A in \cref{fig:teaser}).
In this 2D representation of the solution, we can show its evolution.
Connecting subsequent joint configurations of $s$ through line segments, yields a time curve describing the robot motion.
Connecting the same joint configuration of subsequent intermediate solutions $s_i$, results in a time curve describing the evolution of that particular configuration throughout the optimization. \Cref{fig:pathevolution} illustrates this technique that is integrated in view A in \cref{fig:teaser}.

To combine constraints and solution evolution we adapt the loss landscape technique~\cite{losslandscape} to NLP, which also enables us to observe the behavior of the optimizer as it travels through high-dimensional space (view B in \cref{fig:teaser}).
Using isobands, we can show contours of the objective function and constraints on a 2D plane slice.
The slice plane is spanned by the vector connecting the point of the currently selected optimization step with the minimum $x^\ast$, and a perpendicular vector that is a linear combination of the first two principal vectors of the optimization trajectory $S$.
Selecting three optimization steps aligns the slice to coincide with the corresponding points.
We draw the optimization trajectory into the same plot and encode proximity to the plane as line thickness to determine when the optimizer moves away from it, as it is crucial for judging the degree of correspondence between landscape and trajectory.

For a quick overview of the optimization's descend speed, we employ a view that we call progression speed plot (view E of \cref{fig:teaser}).
This plot uses the metaphor of walking downhill (descending) to the minimum, and shows the remaining distance to travel for each optimization step (where the complete travel distance is normalized to 1).
Task T2 is supported with this view.

Exploration of an optimization is enabled by interaction through selecting or cycling through optimization steps, constraints, or configurations, as well as through zooming and panning in the different views that respond to the selections made.

To select an optimization step, the user can either click into the progression speed and constraint plots, use a slider (on top of the GUI), or select a corresponding entry from an optimization log entry list (H in \cref{fig:teaser}).
The log entries give an overview of the inner workings of the optimization algorithm throughout the process, such as Newton steps taken, line search condition checked, or dual parameter updates done.
We highlight line search backtracking steps in pink, graph query entries correspond to optimization steps.
The naming of these entries stems from the structure of the log, output by the optimizer.
Selecting individual joint configurations (robot time instants) can be done from list G in \cref{fig:teaser}, which results in highlighting the corresponding trajectory in the path evolution view (A).

\subsection{Implementation}
We implemented our visual analytics system as a desktop application in the Java programming language based on the \emph{AWT/Swing} GUI environment.
Libraries employed include \emph{Smile}, \emph{EJML}, \emph{JPlotter}, \emph{jackson}, and \emph{OkHttp}.
The solver runs in a separate software \showchange{(\emph{KOMO}~\cite{komo})}, producing optimization log files that our implementation facilitates to explore interactively.
To obtain samples of the optimization space, we implemented an HTTP-server based interface in the motion planning framework, that runs simultaneously.
We chose this method of data transfer due to its versatility and sustainability for future projects.

\section{Case Study}
\begin{figure*}
	\centering
	\begin{subfigure}{0.32\linewidth}
		\includegraphics[width=\linewidth]{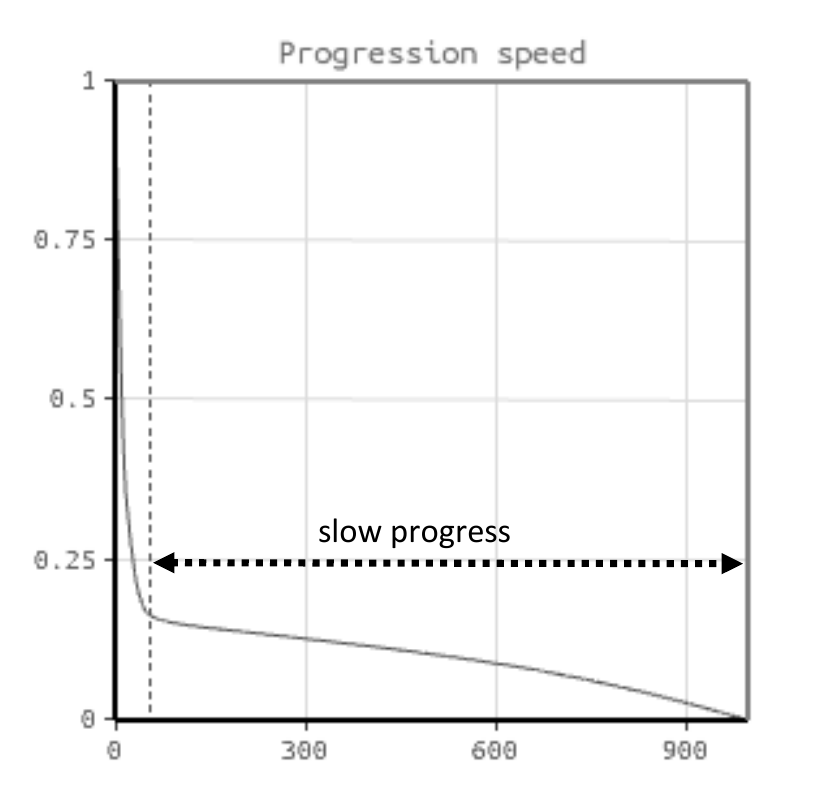}
		\caption{progression overview}
	\end{subfigure}
	\begin{subfigure}{0.32\linewidth}
		\includegraphics[width=\linewidth]{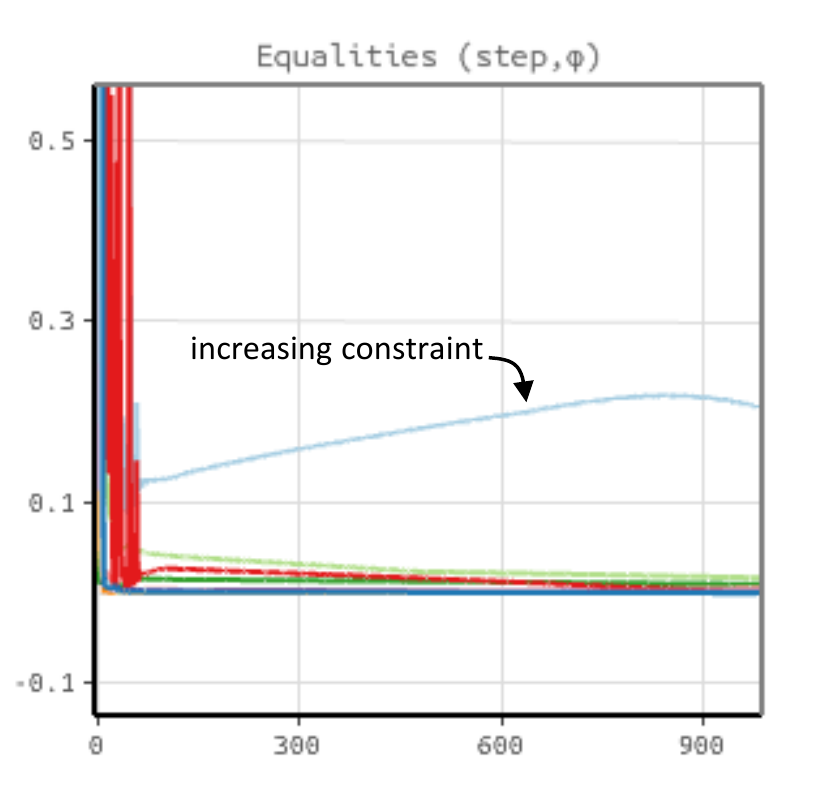}
		\caption{constraint value evolution}
	\end{subfigure}
	\begin{subfigure}{0.32\linewidth}
		\includegraphics[width=\linewidth]{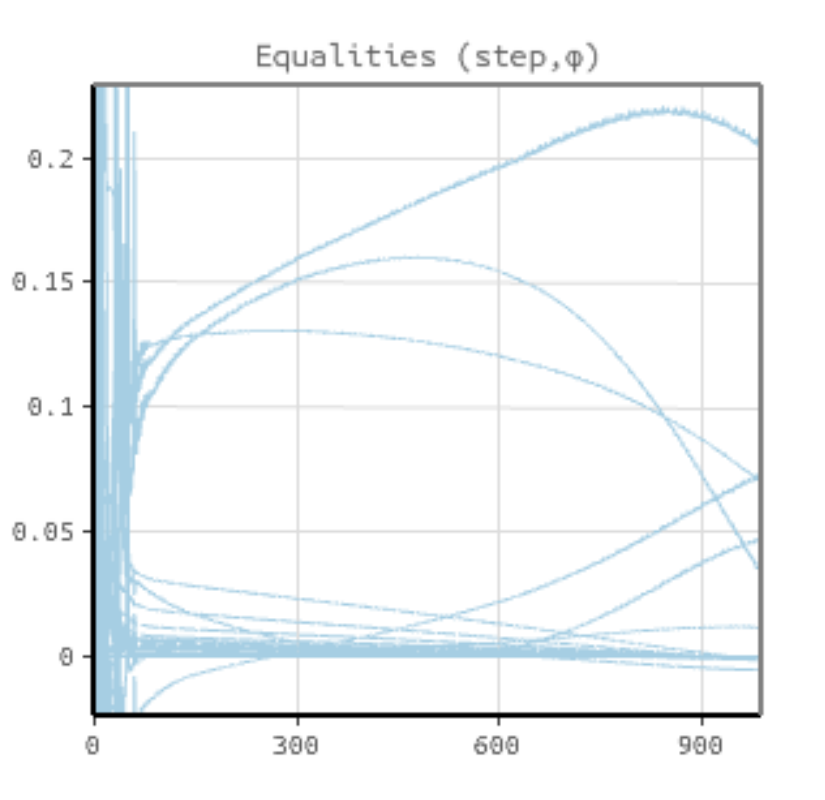}
		\caption{expanded constraint group}
	\end{subfigure}
	\caption{
		Views examined in the first part of our case study.
		We observe slow progress throughout the majority of the optimization (a).
		We find an equality constraint group that increases over the process (b), and on expansion, see that several constraints of the group behave undesirably (c), \showchange{i.e., not converging to zero}.
		%\vspace{-2ex}
	}
	\label{fig:casestudyviews1}
\end{figure*}
To showcase the usefulness of our system we present a case study and insights we could find with respect to a particular motion planning problem.
We analyze an optimization run of a typical motion planning problem concerned with information need Q1.
The robot in this problem is supposed to pick up and throw a ball so that it bounces of the ground and a wall to finally hit a target area.

When taking a look at the corresponding optimization in our visual analytics system, we would like to first get an overview.
We typically start with checking the progression speed plot from which we can see the, in this case, large number of steps taken, which is around 1000~(\cref{fig:casestudyviews1}).
Examining the process' progression behavior (T2), we can observe that the optimizer is leaping forward in the first few iterations and then harshly slows down.
Clicking the plot where we identified the sudden decrease in speed selects the corresponding optimization step (step 57 in this case).
From there on, the optimizer crawls to its final position also hardly accelerating over the remaining iterations.
This behavior is quite suspicious to us since this seems extremely ineffective, so we want to know what is going on.

Next, we take a look at the evolution of constraint values to check for any anomalies there (T3).
Zooming in a little on the equality constraint line chart reveals a constraint that is increasing instead of decreasing during the slow part of the optimization~(blue line in \cref{fig:casestudyviews1}).
The graph shows the maximum absolute value for the constraints sharing the same name, so to identify which constraints exactly are behaving in this way we expand this group of constraints to see the individuals~(\cref{fig:casestudyviews1}).
From the expanded constraints, we see that most of them actually converge to zero as desired, but some of them rise up in the end and one is located far away from zero.

\begin{figure}
	\centering
	\includegraphics[width=0.9\linewidth]{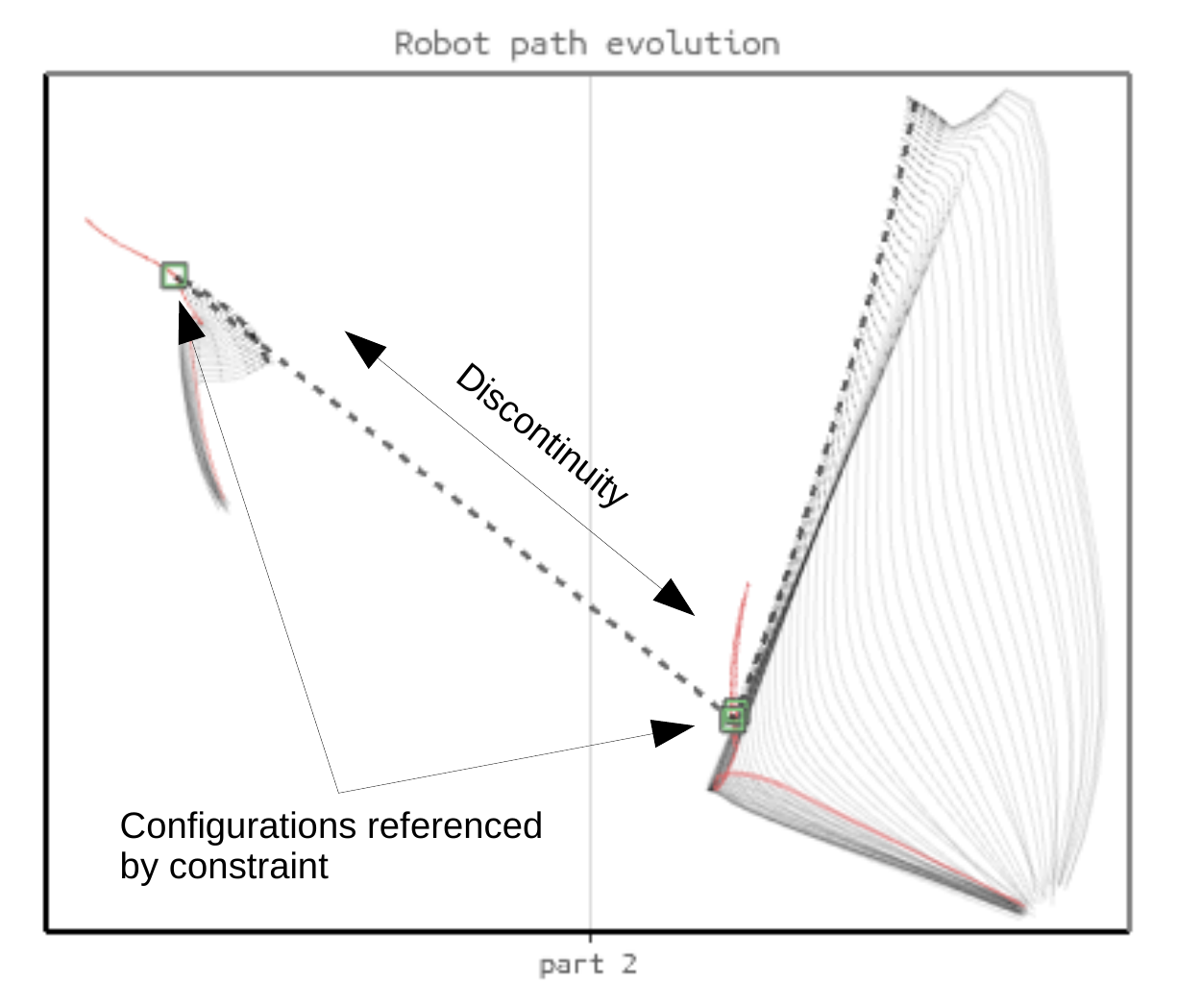}
	\caption{
		Examining the part on the motion path that the misbehaving constraints relate to. We see a discontinuity in the path that corresponds to a change in configuration vector semantics.
		The constraints are related to the moment when the robot releases the ball.
		%\vspace{-3ex}
	}
	\label{fig:pathevocase}
\end{figure}

From the corresponding entries in the constraint tree, we can read off the robot time instants, i.e., the joint configurations to which the strange behaving constraints relate.
Three consecutive configurations in the second part of the motion are related.
This tells us that it is the part where the robot grabs the ball and throws it.
Selecting the entry highlights the trajectories in the path evolution view.
We use this to get an idea of the region on the path that is affected by the constraint~(\cref{fig:pathevocase}).
Examining this part, we recognize a discontinuity in the motion path.
This is due to the semantics of dimensions changing from one configuration to the next, indicating an event on the motion path that requires a different mathematical modeling.
It tells us that the constraint is related to the exact moment of releasing the ball from its grip.
This information could be valuable to the author of the motion problem when wanting to reformulate the problem to achieve better convergence.

\begin{figure*}
	\centering		
	\begin{subfigure}{0.64\linewidth}
		\includegraphics[width=\linewidth]{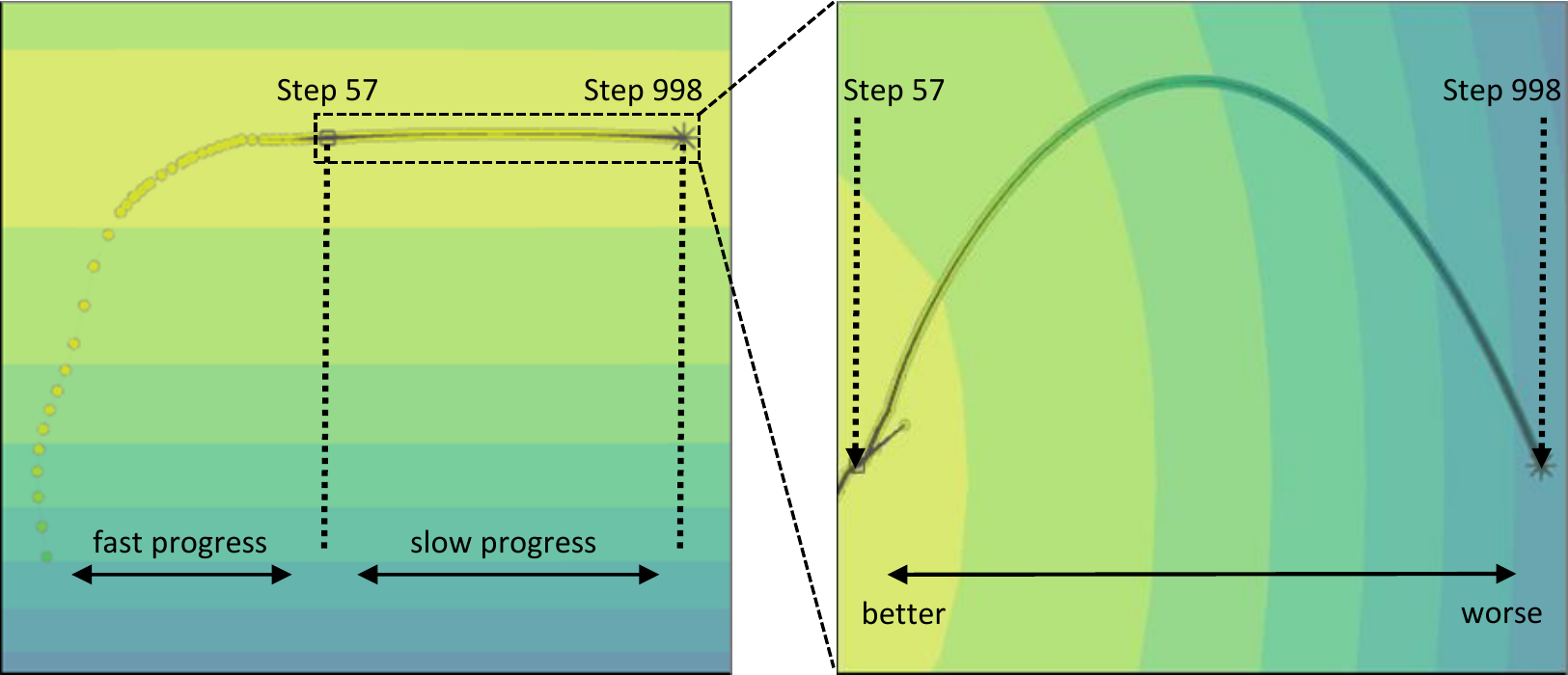}
	\end{subfigure}
	\hspace{1pt}
	\begin{subfigure}{0.3\linewidth}
		\includegraphics[width=\linewidth]{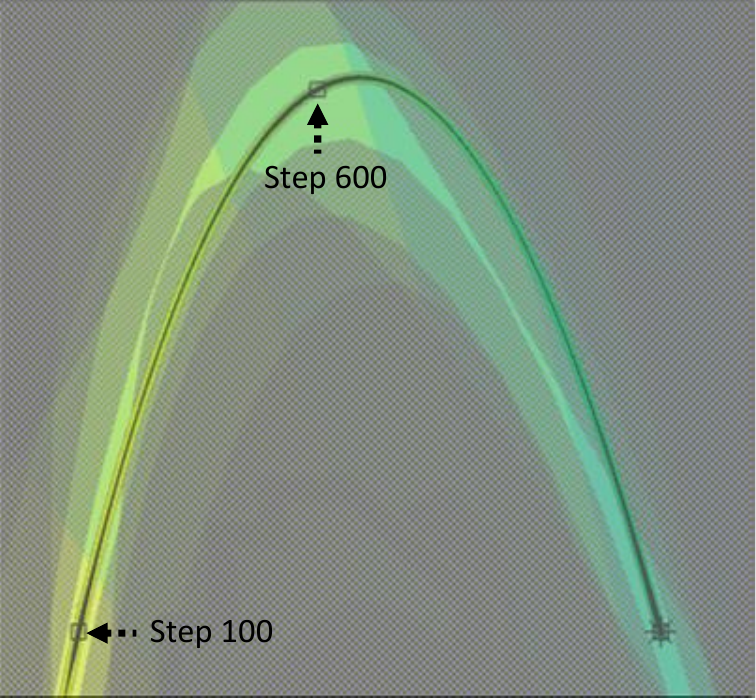}
	\end{subfigure}		
	\caption{
		Optimization landscape view examined in the last part of our case study.
		We zoom-in on the part of the optimization trajectory corresponding to slow progression we identified earlier (left).
		We can observe how the optimizer is climbing uphill in terms of the objective function (middle).
		When displaying the constraint boundaries, we see that the trajectory is moving in a parabola-shaped valley defined by the constraints (right).
		%\vspace{-2ex}
	}
	\label{fig:casestudyviews2}
\end{figure*}

We now want to go back to the original issue of slow progression.
Checking the log table for frequent line searches that could cause slow progression reveals that there are no line searches taking place in the largest portion of the process.
As a next step, we consult the landscape view to take a look at the optimization trajectory~(\cref{fig:casestudyviews2}).
We can observe that the optimizer actually moves away from a relatively better location in terms of its objective function indicated by the colored contours (yellow is less costly than green).
This is not unusual in constraint optimization since the optimizer has to make a compromise to satisfy the constraints.
Our hypothesis is that the suspected constraint forces the optimizer to leave this \enquote{cozy} place.

To test this hypothesis, we enable the display of constraint boundaries in this view (T4).
We also make sure that the projection plane is representative of the slowly progressing part on the trajectory by selecting three points at steps 100, 600, and the final solution~(\cref{fig:casestudyviews2}).
This orientates the plane to coincide with these three points so that we get a reasonable plane for our landscape.
We can observe from this view that the trajectory moves in a parabolic shape and is actually enclosed by boundaries that have a similar shape.
The areas shaded in gray correspond to infeasible regions of the suspected constraints for a certain feasibility threshold that we increased from one to six to be able to see a region at this large scale.
Our interpretation of this is that the optimizer is trying to walk as closely as possible to the hyper surfaces defined by the constraint equations.
The reason it keeps on walking is that it has not yet reached a location where it is close enough to all surfaces to be considered a feasible solution.

Considering the progression speed again and also no line search behavior for the majority of the trajectory, taking so many optimization steps still seems unreasonable.
A sharper adaptive increase in step size was tested afterwards and led to a shortened time to convergence.

Since a single analysis scenario can only show some of the various issues that can occur in optimizations, we like to point the reader to our supplemental video for further impressions.
\begin{printonly}
In the video, we provide two other analysis scenarios while also demonstrating the system's interactive features, which are beyond the scope of this paper. Supplementary materials are available in the online version of this paper.
\end{printonly}
\begin{screenonly}
In the video, we provide two other analysis scenarios while also demonstrating the system's interactive features, which are beyond the scope of this paper~\footnote{\url{https://doi.org/10.18419/darus-1128}}.
\end{screenonly}

\section{Discussion}
In this section, we reflect on the project and report on the lessons learned during our research.
NLP for motion planning is quite complex and comes with many quantities to visualize, such as objective costs, constraints, dual variables, gradient forces, and hypersurfaces --- just to name a few. 
On top of that, it comes in high-dimensional flavors and two notions of time (optimization time and robot time).
We chose to focus on the evolutionary aspect, in terms of optimization time, to be able to follow the algorithm along its process.
This choice turned out to work well for the assessment of long lasting optimizations and causes for that (Q1).

The use of line charts to examine convergence behavior of constraints or loss is a widely used practice in optimization.
They serve well as an entry point to optimization assessment, but experts want to be able to reason beyond the scope of such charts in order to understand what is actually happening.
Leveraging linked views to allow for a more thorough exploration of the process worked well in our case.

During the course of this design study, we noticed that slow progression can be connected to ill conditioning and that long lasting optimizations often fail to produce feasible results.
To assess optimizations with infeasible solutions (Q2), we also wanted to be able to reason about the behavior of the algorithm, analyze why it goes which way, and ideally see when it takes a \enquote{wrong} turn.
Through the loss landscape visualization, we were able to provide means to achieve this.
However, this technique is the most effective if the optimizer only moves in a low-dimensional subspace of the complete optimization space.
This is not always the case and plane orientation becomes the deciding factor to effectively use this view.

To understand abstract high-dimensional points of optimization space, we chose a generic solution to display such time-involving intermediate solutions (motion paths in our case) by means of time curves.
This works to some degree, however, experts would value a more concrete representation for their domain specific application, e.g., an animation of the robot performing the motion to better connect the abstract and physical world.
\showchange{We expect that integrating the animation capabilities of the motion planning framework into our system will greatly improve optimization space exploration in future work.}

\section{Conclusion}
We presented a visual analytics system for the assessment of optimization processes of nonlinear constraint problems for robot motion planning.
Except for the robot path evolution view, displaying the evolution of the motion path, the system provides domain-agnostic views for the analysis of nonlinear programs.
Leveraging a loss landscape visualization technique, the system allows users to have a look at the optimization trajectory and surrounding constraints even for the very high-dimensional problems we are facing in the domain.
However, we found that the choice of projection plane is crucial for this technique to be effective.
Better plane orientation could be explored in future research.
The system's capabilities were showcased in a case study where we analyzed optimization runs of our collaborators' NLPs.
Together with domain experts, we could confirm that we are able to gain new insights into optimizer behavior, detect flaws in the optimization process, and come up with issue resolving strategies by exploration and analysis with our system.
Since most views are applicable to any nonlinear program, we think that our approach to nonlinear constraint optimization visualization will generalize to other problems.
In future work, we plan to investigate respective applications and evaluate our approach in more depth.

\begin{acks}
The research has been supported by the Deutsche Forschungsgemeinschaft (DFG, German Research Foundation) under Germany's Excellence Strategy -- EXC 2120/1 -- 390831618, and the DAAD.
\end{acks}

\bibliographystyle{ACM-Reference-Format}
\bibliography{template}


\begin{thebibliography}{46}

%%% the Web address.

\ifx \showCODEN    \undefined \def \showCODEN     #1{\unskip}     \fi
\ifx \showDOI      \undefined \def \showDOI       #1{#1}\fi
\ifx \showISBNx    \undefined \def \showISBNx     #1{\unskip}     \fi
\ifx \showISBNxiii \undefined \def \showISBNxiii  #1{\unskip}     \fi
\ifx \showISSN     \undefined \def \showISSN      #1{\unskip}     \fi
\ifx \showLCCN     \undefined \def \showLCCN      #1{\unskip}     \fi
\ifx \shownote     \undefined \def \shownote      #1{#1}          \fi
\ifx \showarticletitle \undefined \def \showarticletitle #1{#1}   \fi
\ifx \showURL      \undefined \def \showURL       {\relax}        \fi
% invisible to TeX
\providecommand\bibfield[2]{#2}
\providecommand\bibinfo[2]{#2}
\providecommand\natexlab[1]{#1}
\providecommand\showeprint[2][]{arXiv:#2}

\bibitem[\protect\citeauthoryear{{Aigner}, {Miksch}, {Schumann}, and
  {Tominski}}{{Aigner} et~al\mbox{.}}{2011}]%
        {aigner2011visualization}
\bibfield{author}{\bibinfo{person}{W. {Aigner}}, \bibinfo{person}{S. {Miksch}},
  \bibinfo{person}{H. {Schumann}}, {and} \bibinfo{person}{C. {Tominski}}.}
  \bibinfo{year}{2011}\natexlab{}.
\newblock \bibinfo{booktitle}{\emph{Visualization of Time-Oriented Data}}.
\newblock \bibinfo{publisher}{Springer Science \& Business Media}.
\newblock


\bibitem[\protect\citeauthoryear{Andrews}{Andrews}{1972}]%
        {andrews1972plots}
\bibfield{author}{\bibinfo{person}{David~F Andrews}.}
  \bibinfo{year}{1972}\natexlab{}.
\newblock \showarticletitle{Plots of high-dimensional data}.
\newblock \bibinfo{journal}{\emph{Biometrics}} \bibinfo{volume}{18},
  \bibinfo{number}{1} (\bibinfo{year}{1972}), \bibinfo{pages}{125--136}.
\newblock


\bibitem[\protect\citeauthoryear{Androulakis and Vrahatis}{Androulakis and
  Vrahatis}{1996}]%
        {ANDROULAKIS199641}
\bibfield{author}{\bibinfo{person}{G.S. Androulakis} {and}
  \bibinfo{person}{M.N. Vrahatis}.} \bibinfo{year}{1996}\natexlab{}.
\newblock \showarticletitle{{OPTAC:} a portable software package for analyzing
  and comparing optimization methods by visualization}.
\newblock \bibinfo{journal}{\emph{J. Comput. Appl. Math.}}
  \bibinfo{volume}{72}, \bibinfo{number}{1} (\bibinfo{year}{1996}),
  \bibinfo{pages}{41--62}.
\newblock


\bibitem[\protect\citeauthoryear{Bach, Dragicevic, Archambault, Hurter, and
  Carpendale}{Bach et~al\mbox{.}}{2014}]%
        {spacetimecubeoperations}
\bibfield{author}{\bibinfo{person}{B. Bach}, \bibinfo{person}{P. Dragicevic},
  \bibinfo{person}{D. Archambault}, \bibinfo{person}{C. Hurter}, {and}
  \bibinfo{person}{S. Carpendale}.} \bibinfo{year}{2014}\natexlab{}.
\newblock \showarticletitle{A Review of Temporal Data Visualizations Based on
  Space-Time Cube Operations}. In \bibinfo{booktitle}{\emph{EuroVis -- STARs}}.
  \bibinfo{pages}{23--41}.
\newblock
\showISBNx{978-3-03868-028-4}


\bibitem[\protect\citeauthoryear{{Bach}, {Henry-Riche}, {Dwyer}, {Madhyastha},
  {Fekete}, and {Grabowski}}{{Bach} et~al\mbox{.}}{2015}]%
        {bach2015small}
\bibfield{author}{\bibinfo{person}{B. {Bach}}, \bibinfo{person}{N.
  {Henry-Riche}}, \bibinfo{person}{T. {Dwyer}}, \bibinfo{person}{T.
  {Madhyastha}}, \bibinfo{person}{J-D {Fekete}}, {and} \bibinfo{person}{T
  {Grabowski}}.} \bibinfo{year}{2015}\natexlab{}.
\newblock \showarticletitle{{Small MultiPiles:} Piling time to explore temporal
  patterns in dynamic networks}.
\newblock \bibinfo{journal}{\emph{Computer Graphics Forum}}
  \bibinfo{volume}{34}, \bibinfo{number}{3} (\bibinfo{year}{2015}),
  \bibinfo{pages}{31--40}.
\newblock


\bibitem[\protect\citeauthoryear{{Bach}, {Shi}, {Heulot}, {Madhyastha},
  {Grabowski}, and {Dragicevic}}{{Bach} et~al\mbox{.}}{2016}]%
        {timecurves}
\bibfield{author}{\bibinfo{person}{B. {Bach}}, \bibinfo{person}{C. {Shi}},
  \bibinfo{person}{N. {Heulot}}, \bibinfo{person}{T. {Madhyastha}},
  \bibinfo{person}{T. {Grabowski}}, {and} \bibinfo{person}{P. {Dragicevic}}.}
  \bibinfo{year}{2016}\natexlab{}.
\newblock \showarticletitle{{Time Curves:} Folding Time to Visualize Patterns
  of Temporal Evolution in Data}.
\newblock \bibinfo{journal}{\emph{IEEE Transactions on Visualization and
  Computer Graphics}} \bibinfo{volume}{22}, \bibinfo{number}{1}
  (\bibinfo{year}{2016}), \bibinfo{pages}{559--568}.
\newblock


\bibitem[\protect\citeauthoryear{Carro and Hermenegildo}{Carro and
  Hermenegildo}{2000a}]%
        {carro2000VIFIDtool}
\bibfield{author}{\bibinfo{person}{Manuel Carro} {and} \bibinfo{person}{Manuel
  Hermenegildo}.} \bibinfo{year}{2000}\natexlab{a}.
\newblock \showarticletitle{Tools for constraint visualisation: The
  {VIFID/TRIFID} tool}.
\newblock In \bibinfo{booktitle}{\emph{Analysis and Visualization Tools for
  Constraint Programming}}. \bibinfo{publisher}{Springer},
  \bibinfo{pages}{253--272}.
\newblock


\bibitem[\protect\citeauthoryear{Carro and Hermenegildo}{Carro and
  Hermenegildo}{2000b}]%
        {carro2000apttool}
\bibfield{author}{\bibinfo{person}{Manuel Carro} {and} \bibinfo{person}{Manuel
  Hermenegildo}.} \bibinfo{year}{2000}\natexlab{b}.
\newblock \showarticletitle{Tools for search-tree visualisation: The apt tool}.
\newblock In \bibinfo{booktitle}{\emph{Analysis and Visualization Tools for
  Constraint Programming}}. \bibinfo{publisher}{Springer},
  \bibinfo{pages}{237--252}.
\newblock


\bibitem[\protect\citeauthoryear{{Charalambos} and {Izquierdo}}{{Charalambos}
  and {Izquierdo}}{2001}]%
        {LPviz}
\bibfield{author}{\bibinfo{person}{J.~P. {Charalambos}} {and}
  \bibinfo{person}{E. {Izquierdo}}.} \bibinfo{year}{2001}\natexlab{}.
\newblock \showarticletitle{Linear programming concept visualization}. In
  \bibinfo{booktitle}{\emph{Proceedings Fifth International Conference on
  Information Visualisation}}. \bibinfo{pages}{529--535}.
\newblock


\bibitem[\protect\citeauthoryear{Chatterjee, Das, and Bhattacharya}{Chatterjee
  et~al\mbox{.}}{1993}]%
        {LPparallelcoords}
\bibfield{author}{\bibinfo{person}{A Chatterjee}, \bibinfo{person}{P.P Das},
  {and} \bibinfo{person}{S Bhattacharya}.} \bibinfo{year}{1993}\natexlab{}.
\newblock \showarticletitle{Visualization in linear programming using parallel
  coordinates}.
\newblock \bibinfo{journal}{\emph{Pattern Recognition}} \bibinfo{volume}{26},
  \bibinfo{number}{11} (\bibinfo{year}{1993}), \bibinfo{pages}{1725 -- 1736}.
\newblock


\bibitem[\protect\citeauthoryear{{Chen} and {Hwang}}{{Chen} and
  {Hwang}}{1998}]%
        {SANDROS}
\bibfield{author}{\bibinfo{person}{P.~C. {Chen}} {and} \bibinfo{person}{Y.~K.
  {Hwang}}.} \bibinfo{year}{1998}\natexlab{}.
\newblock \showarticletitle{SANDROS: a dynamic graph search algorithm for
  motion planning}.
\newblock \bibinfo{journal}{\emph{IEEE Transactions on Robotics and
  Automation}} \bibinfo{volume}{14}, \bibinfo{number}{3}
  (\bibinfo{year}{1998}), \bibinfo{pages}{390--403}.
\newblock


\bibitem[\protect\citeauthoryear{Chernoff}{Chernoff}{1973}]%
        {chernoff1973use}
\bibfield{author}{\bibinfo{person}{Herman Chernoff}.}
  \bibinfo{year}{1973}\natexlab{}.
\newblock \showarticletitle{The use of faces to represent points in
  k-dimensional space graphically}.
\newblock \bibinfo{journal}{\emph{J. Amer. Statist. Assoc.}}
  \bibinfo{volume}{68}, \bibinfo{number}{342} (\bibinfo{year}{1973}),
  \bibinfo{pages}{361--368}.
\newblock


\bibitem[\protect\citeauthoryear{Chinneck}{Chinneck}{2006}]%
        {chinneck2006practical}
\bibfield{author}{\bibinfo{person}{John~W Chinneck}.}
  \bibinfo{year}{2006}\natexlab{}.
\newblock \showarticletitle{Practical optimization: a gentle introduction}.
\newblock \bibinfo{journal}{\emph{Systems and Computer Engineering, Carleton
  University, Ottawa}} (\bibinfo{year}{2006}).
\newblock


\bibitem[\protect\citeauthoryear{{Ghoniem}, {Cambazard}, {Fekete}, and
  {Jussien}}{{Ghoniem} et~al\mbox{.}}{2005}]%
        {Ghoniem2005Peeking}
\bibfield{author}{\bibinfo{person}{M. {Ghoniem}}, \bibinfo{person}{H.
  {Cambazard}}, \bibinfo{person}{J-D {Fekete}}, {and} \bibinfo{person}{N.
  {Jussien}}.} \bibinfo{year}{2005}\natexlab{}.
\newblock \showarticletitle{Peeking in Solver Strategies Using Explanations
  Visualization of Dynamic Graphs for Constraint Programming}. In
  \bibinfo{booktitle}{\emph{Proceedings of the 2005 ACM Symposium on Software
  Visualization}}. \bibinfo{pages}{27--36}.
\newblock
\showISBNx{1595930736}


\bibitem[\protect\citeauthoryear{{Ghoniem}, {Jussien}, and {Fekete}}{{Ghoniem}
  et~al\mbox{.}}{2004}]%
        {ghoniem2004visexp}
\bibfield{author}{\bibinfo{person}{M. {Ghoniem}}, \bibinfo{person}{N.
  {Jussien}}, {and} \bibinfo{person}{J-D {Fekete}}.}
  \bibinfo{year}{2004}\natexlab{}.
\newblock \showarticletitle{Visexp: visualizing constraint solver dynamics
  using explanations}. In \bibinfo{booktitle}{\emph{FLAIRS'04: Seventeenth
  international Florida Artificial Intelligence Research Society conference}}.
  \bibinfo{pages}{263--268}.
\newblock


\bibitem[\protect\citeauthoryear{Goodfellow and Vinyals}{Goodfellow and
  Vinyals}{2015}]%
        {characterizingnn}
\bibfield{author}{\bibinfo{person}{Ian~J. Goodfellow} {and}
  \bibinfo{person}{Oriol Vinyals}.} \bibinfo{year}{2015}\natexlab{}.
\newblock \showarticletitle{Qualitatively characterizing neural network
  optimization problems}. In \bibinfo{booktitle}{\emph{3rd International
  Conference on Learning Representations, {ICLR}}}.
\newblock


\bibitem[\protect\citeauthoryear{{Goodwin}, {Mears}, {Dwyer}, {de la Banda},
  {Tack}, and {Wallace}}{{Goodwin} et~al\mbox{.}}{2017}]%
        {constraintprogrammingstudy}
\bibfield{author}{\bibinfo{person}{S. {Goodwin}}, \bibinfo{person}{C. {Mears}},
  \bibinfo{person}{T. {Dwyer}}, \bibinfo{person}{M.~G. {de la Banda}},
  \bibinfo{person}{G. {Tack}}, {and} \bibinfo{person}{M. {Wallace}}.}
  \bibinfo{year}{2017}\natexlab{}.
\newblock \showarticletitle{What do Constraint Programming Users Want to See?
  {E}xploring the Role of Visualisation in Profiling of Models and Search}.
\newblock \bibinfo{journal}{\emph{IEEE Transactions on Visualization and
  Computer Graphics}} \bibinfo{volume}{23}, \bibinfo{number}{1}
  (\bibinfo{year}{2017}), \bibinfo{pages}{281--290}.
\newblock


\bibitem[\protect\citeauthoryear{Gruendl, Riehmann, Pausch, and
  Froehlich}{Gruendl et~al\mbox{.}}{2016}]%
        {gruendl2016time}
\bibfield{author}{\bibinfo{person}{Henning Gruendl}, \bibinfo{person}{Patrick
  Riehmann}, \bibinfo{person}{Yves Pausch}, {and} \bibinfo{person}{Bernd
  Froehlich}.} \bibinfo{year}{2016}\natexlab{}.
\newblock \showarticletitle{Time-Series Plots Integrated in
  Parallel-Coordinates Displays}.
\newblock \bibinfo{journal}{\emph{Computer Graphics Forum}}
  \bibinfo{volume}{35}, \bibinfo{number}{3} (\bibinfo{year}{2016}),
  \bibinfo{pages}{321--330}.
\newblock


\bibitem[\protect\citeauthoryear{Heipcke}{Heipcke}{1999}]%
        {Heipcke1999}
\bibfield{author}{\bibinfo{person}{S Heipcke}.}
  \bibinfo{year}{1999}\natexlab{}.
\newblock \showarticletitle{Comparing constraint programming and mathematical
  programming approaches to discrete optimisation - the change problem}.
\newblock \bibinfo{journal}{\emph{Journal of the Operational Research Society}}
  \bibinfo{volume}{50}, \bibinfo{number}{6} (\bibinfo{year}{1999}),
  \bibinfo{pages}{581--595}.
\newblock


\bibitem[\protect\citeauthoryear{Inselberg}{Inselberg}{1985}]%
        {inselberg1985plane}
\bibfield{author}{\bibinfo{person}{Alfred Inselberg}.}
  \bibinfo{year}{1985}\natexlab{}.
\newblock \showarticletitle{The plane with parallel coordinates}.
\newblock \bibinfo{journal}{\emph{The Visual Computer}} \bibinfo{volume}{1},
  \bibinfo{number}{2} (\bibinfo{year}{1985}), \bibinfo{pages}{69--91}.
\newblock


\bibitem[\protect\citeauthoryear{{J\"ackle}, {Fischer}, {Schreck}, and
  {Keim}}{{J\"ackle} et~al\mbox{.}}{2016}]%
        {temporalmds}
\bibfield{author}{\bibinfo{person}{D. {J\"ackle}}, \bibinfo{person}{F.
  {Fischer}}, \bibinfo{person}{T. {Schreck}}, {and} \bibinfo{person}{D.~A.
  {Keim}}.} \bibinfo{year}{2016}\natexlab{}.
\newblock \showarticletitle{Temporal {MDS} Plots for Analysis of Multivariate
  Data}.
\newblock \bibinfo{journal}{\emph{IEEE Transactions on Visualization and
  Computer Graphics}} \bibinfo{volume}{22}, \bibinfo{number}{1}
  (\bibinfo{year}{2016}), \bibinfo{pages}{141--150}.
\newblock
\showISSN{1077-2626}


\bibitem[\protect\citeauthoryear{Kandogan}{Kandogan}{2000}]%
        {kandogan2000star}
\bibfield{author}{\bibinfo{person}{Eser Kandogan}.}
  \bibinfo{year}{2000}\natexlab{}.
\newblock \showarticletitle{Star coordinates: A multi-dimensional visualization
  technique with uniform treatment of dimensions}. In
  \bibinfo{booktitle}{\emph{Proceedings of the IEEE Information Visualization
  Symposium}}.
\newblock


\bibitem[\protect\citeauthoryear{{Kavraki} and {Latombe}}{{Kavraki} and
  {Latombe}}{1994}]%
        {Kavraki1994}
\bibfield{author}{\bibinfo{person}{L. {Kavraki}} {and} \bibinfo{person}{J.
  {Latombe}}.} \bibinfo{year}{1994}\natexlab{}.
\newblock \showarticletitle{Randomized preprocessing of configuration for fast
  path planning}. In \bibinfo{booktitle}{\emph{Proceedings of the IEEE
  International Conference on Robotics and Automation}}.
  \bibinfo{pages}{2138--2145, vol. 3}.
\newblock


\bibitem[\protect\citeauthoryear{{Kavraki}, {Svestka}, {Latombe}, and
  {Overmars}}{{Kavraki} et~al\mbox{.}}{1996}]%
        {Kavraki1996}
\bibfield{author}{\bibinfo{person}{L.~E. {Kavraki}}, \bibinfo{person}{P.
  {Svestka}}, \bibinfo{person}{J. {Latombe}}, {and} \bibinfo{person}{M.~H.
  {Overmars}}.} \bibinfo{year}{1996}\natexlab{}.
\newblock \showarticletitle{Probabilistic roadmaps for path planning in
  high-dimensional configuration spaces}.
\newblock \bibinfo{journal}{\emph{IEEE Transactions on Robotics and
  Automation}} \bibinfo{volume}{12}, \bibinfo{number}{4}
  (\bibinfo{year}{1996}), \bibinfo{pages}{566--580}.
\newblock


\bibitem[\protect\citeauthoryear{Kondo}{Kondo}{1991}]%
        {KONDO1991}
\bibfield{author}{\bibinfo{person}{K Kondo}.} \bibinfo{year}{1991}\natexlab{}.
\newblock \showarticletitle{Motion planning with six degrees of freedom by
  multistrategic bidirectional heuristic free-space enumeration}.
\newblock \bibinfo{journal}{\emph{IEEE Transactions on Robotics and
  Automation}} \bibinfo{volume}{7}, \bibinfo{number}{3} (\bibinfo{year}{1991}),
  \bibinfo{pages}{267--277}.
\newblock
\showISSN{1042-296X}


\bibitem[\protect\citeauthoryear{Kuhn and Tucker}{Kuhn and Tucker}{2014}]%
        {kuhntuckernlp}
\bibfield{author}{\bibinfo{person}{Harold~W. Kuhn} {and}
  \bibinfo{person}{Albert~W. Tucker}.} \bibinfo{year}{2014}\natexlab{}.
\newblock \bibinfo{booktitle}{\emph{Nonlinear Programming}}.
\newblock \bibinfo{publisher}{Springer Basel}, \bibinfo{address}{Basel},
  \bibinfo{pages}{247--258}.
\newblock


\bibitem[\protect\citeauthoryear{Latombe}{Latombe}{2012}]%
        {latombe2012robot}
\bibfield{author}{\bibinfo{person}{Jean-Claude Latombe}.}
  \bibinfo{year}{2012}\natexlab{}.
\newblock \bibinfo{booktitle}{\emph{Robot Motion Planning}}.
  Vol.~\bibinfo{volume}{124}.
\newblock \bibinfo{publisher}{Springer Science \& Business Media}.
\newblock


\bibitem[\protect\citeauthoryear{LaValle}{LaValle}{2006}]%
        {lavalle2006planning}
\bibfield{author}{\bibinfo{person}{Steven~M LaValle}.}
  \bibinfo{year}{2006}\natexlab{}.
\newblock \bibinfo{booktitle}{\emph{Planning Algorithms}}.
\newblock \bibinfo{publisher}{Cambridge University Press}.
\newblock


\bibitem[\protect\citeauthoryear{Li, Xu, Taylor, Studer, and Goldstein}{Li
  et~al\mbox{.}}{2018}]%
        {losslandscape}
\bibfield{author}{\bibinfo{person}{Hao Li}, \bibinfo{person}{Zheng Xu},
  \bibinfo{person}{Gavin Taylor}, \bibinfo{person}{Christoph Studer}, {and}
  \bibinfo{person}{Tom Goldstein}.} \bibinfo{year}{2018}\natexlab{}.
\newblock \showarticletitle{Visualizing the Loss Landscape of Neural Nets}.
\newblock In \bibinfo{booktitle}{\emph{Advances in Neural Information
  Processing Systems 31}}. \bibinfo{pages}{6389--6399}.
\newblock


\bibitem[\protect\citeauthoryear{{Liu}, {Maljovec}, {Wang}, {Bremer}, and
  {Pascucci}}{{Liu} et~al\mbox{.}}{2017}]%
        {Liu2017}
\bibfield{author}{\bibinfo{person}{S. {Liu}}, \bibinfo{person}{D. {Maljovec}},
  \bibinfo{person}{B. {Wang}}, \bibinfo{person}{P. {Bremer}}, {and}
  \bibinfo{person}{V. {Pascucci}}.} \bibinfo{year}{2017}\natexlab{}.
\newblock \showarticletitle{Visualizing High-Dimensional Data: Advances in the
  Past Decade}.
\newblock \bibinfo{journal}{\emph{IEEE Transactions on Visualization and
  Computer Graphics}} \bibinfo{volume}{23}, \bibinfo{number}{3}
  (\bibinfo{year}{2017}), \bibinfo{pages}{1249--1268}.
\newblock


\bibitem[\protect\citeauthoryear{{Messac} and {Chen}}{{Messac} and
  {Chen}}{2000}]%
        {MESSAC2000}
\bibfield{author}{\bibinfo{person}{A. {Messac}} {and} \bibinfo{person}{X.
  {Chen}}.} \bibinfo{year}{2000}\natexlab{}.
\newblock \showarticletitle{Visualizing the optimization process in real-time
  using physical programming}.
\newblock \bibinfo{journal}{\emph{Engineering Optimization}}
  \bibinfo{volume}{32}, \bibinfo{number}{6} (\bibinfo{year}{2000}),
  \bibinfo{pages}{721--747}.
\newblock


\bibitem[\protect\citeauthoryear{{Noirhomme-Fraiture}}{{Noirhomme-Fraiture}}{2002}]%
        {noirhomme2002visualization}
\bibfield{author}{\bibinfo{person}{M. {Noirhomme-Fraiture}}.}
  \bibinfo{year}{2002}\natexlab{}.
\newblock \showarticletitle{Visualization of large data sets: the zoom star
  solution}.
\newblock \bibinfo{journal}{\emph{International Electronic Journal of Symbolic
  Data Analysis}} (\bibinfo{year}{2002}), \bibinfo{pages}{26--39}.
\newblock


\bibitem[\protect\citeauthoryear{{Nonato} and {Aupetit}}{{Nonato} and
  {Aupetit}}{2018}]%
        {nonato2018multidimensional}
\bibfield{author}{\bibinfo{person}{L. {Nonato}} {and} \bibinfo{person}{A.
  {Aupetit}}.} \bibinfo{year}{2018}\natexlab{}.
\newblock \showarticletitle{Multidimensional projection for visual analytics:
  Linking techniques with distortions, tasks, and layout enrichment}.
\newblock \bibinfo{journal}{\emph{IEEE Transactions on Visualization and
  Computer Graphics}} \bibinfo{volume}{25}, \bibinfo{number}{8}
  (\bibinfo{year}{2018}), \bibinfo{pages}{2650--2673}.
\newblock


\bibitem[\protect\citeauthoryear{{Pu} and {Lalanne}}{{Pu} and
  {Lalanne}}{2000}]%
        {Pu2000}
\bibfield{author}{\bibinfo{person}{P. {Pu}} {and} \bibinfo{person}{D.
  {Lalanne}}.} \bibinfo{year}{2000}\natexlab{}.
\newblock \showarticletitle{Interactive problem solving via algorithm
  visualization}. In \bibinfo{booktitle}{\emph{IEEE Symposium on Information
  Visualization}}. \bibinfo{pages}{145--153}.
\newblock


\bibitem[\protect\citeauthoryear{{Sedlmair}, {Meyer}, and {Munzner}}{{Sedlmair}
  et~al\mbox{.}}{2012}]%
        {sedlmair}
\bibfield{author}{\bibinfo{person}{M. {Sedlmair}}, \bibinfo{person}{M.
  {Meyer}}, {and} \bibinfo{person}{T. {Munzner}}.}
  \bibinfo{year}{2012}\natexlab{}.
\newblock \showarticletitle{Design Study Methodology: Reflections from the
  Trenches and the Stacks}.
\newblock \bibinfo{journal}{\emph{IEEE Transactions on Visualization and
  Computer Graphics}} \bibinfo{volume}{18}, \bibinfo{number}{12}
  (\bibinfo{year}{2012}), \bibinfo{pages}{2431--2440}.
\newblock


\bibitem[\protect\citeauthoryear{Shishmarev, Mears, Tack, and
  De~La~Banda}{Shishmarev et~al\mbox{.}}{2016}]%
        {shishmarev2016visual}
\bibfield{author}{\bibinfo{person}{Maxim Shishmarev},
  \bibinfo{person}{Christopher Mears}, \bibinfo{person}{Guido Tack}, {and}
  \bibinfo{person}{Maria~Garcia De~La~Banda}.} \bibinfo{year}{2016}\natexlab{}.
\newblock \showarticletitle{Visual search tree profiling}.
\newblock \bibinfo{journal}{\emph{Constraints}} \bibinfo{volume}{21},
  \bibinfo{number}{1} (\bibinfo{year}{2016}), \bibinfo{pages}{77--94}.
\newblock


\bibitem[\protect\citeauthoryear{Simonis, Davern, Feldman, Mehta, Quesada, and
  Carlsson}{Simonis et~al\mbox{.}}{2010}]%
        {cpviz}
\bibfield{author}{\bibinfo{person}{Helmut Simonis}, \bibinfo{person}{Paul
  Davern}, \bibinfo{person}{Jacob Feldman}, \bibinfo{person}{Deepak Mehta},
  \bibinfo{person}{Luis Quesada}, {and} \bibinfo{person}{Mats Carlsson}.}
  \bibinfo{year}{2010}\natexlab{}.
\newblock \showarticletitle{A Generic Visualization Platform for {CP}}. In
  \bibinfo{booktitle}{\emph{International Conference on Principles and Practice
  of Constraint Programming}}. \bibinfo{pages}{460--474}.
\newblock


\bibitem[\protect\citeauthoryear{{Tominski}, {Abello}, and
  {Schumann}}{{Tominski} et~al\mbox{.}}{2004}]%
        {tominski2004axes}
\bibfield{author}{\bibinfo{person}{C. {Tominski}}, \bibinfo{person}{J.
  {Abello}}, {and} \bibinfo{person}{H. {Schumann}}.}
  \bibinfo{year}{2004}\natexlab{}.
\newblock \showarticletitle{Axes-based visualizations with radial layouts}. In
  \bibinfo{booktitle}{\emph{Proceedings of the 2004 ACM Symposium on Applied
  Computing}}. \bibinfo{pages}{1242--1247}.
\newblock


\bibitem[\protect\citeauthoryear{Torsney-Weir, M\"oller, Sedlmair, and
  Kirby}{Torsney-Weir et~al\mbox{.}}{2018}]%
        {hyperslice}
\bibfield{author}{\bibinfo{person}{T. Torsney-Weir}, \bibinfo{person}{T.
  M\"oller}, \bibinfo{person}{M. Sedlmair}, {and} \bibinfo{person}{R.~M.
  Kirby}.} \bibinfo{year}{2018}\natexlab{}.
\newblock \showarticletitle{Hypersliceplorer: Interactive visualization of
  shapes in multiple dimensions}.
\newblock \bibinfo{journal}{\emph{Computer Graphics Forum}}
  \bibinfo{volume}{37}, \bibinfo{number}{3} (\bibinfo{year}{2018}),
  \bibinfo{pages}{229--240}.
\newblock


\bibitem[\protect\citeauthoryear{Toussaint}{Toussaint}{2014}]%
        {komo}
\bibfield{author}{\bibinfo{person}{Marc Toussaint}.}
  \bibinfo{year}{2014}\natexlab{}.
\newblock \showarticletitle{Newton methods for k-order {Markov} Constrained
  Motion Problems}.
\newblock \bibinfo{journal}{\emph{CoRR}}  \bibinfo{volume}{abs/1407.0414}
  (\bibinfo{year}{2014}).
\newblock


\bibitem[\protect\citeauthoryear{Toussaint}{Toussaint}{2015}]%
        {toussaintmotionplanning}
\bibfield{author}{\bibinfo{person}{Marc Toussaint}.}
  \bibinfo{year}{2015}\natexlab{}.
\newblock \showarticletitle{Logic-Geometric Programming: An Optimization-Based
  Approach to Combined Task and Motion Planning}. In
  \bibinfo{booktitle}{\emph{Proceedings of the 24th International Conference on
  Artificial Intelligence}}. \bibinfo{publisher}{AAAI Press},
  \bibinfo{pages}{1930--1936}.
\newblock


\bibitem[\protect\citeauthoryear{Tu{\v{s}}ar and Filipi{\v{c}}}{Tu{\v{s}}ar and
  Filipi{\v{c}}}{2014}]%
        {tuvsar2014visualization}
\bibfield{author}{\bibinfo{person}{Tea Tu{\v{s}}ar} {and}
  \bibinfo{person}{Bogdan Filipi{\v{c}}}.} \bibinfo{year}{2014}\natexlab{}.
\newblock \showarticletitle{Visualization of {Pareto} front approximations in
  evolutionary multiobjective optimization: A critical review and the
  prosection method}.
\newblock \bibinfo{journal}{\emph{IEEE Transactions on Evolutionary
  Computation}} \bibinfo{volume}{19}, \bibinfo{number}{2}
  (\bibinfo{year}{2014}), \bibinfo{pages}{225--245}.
\newblock


\bibitem[\protect\citeauthoryear{van~den Elzen, {Holten}, {Blaas}, and {van
  Wijk}}{van~den Elzen et~al\mbox{.}}{2016}]%
        {snapshotstopoints}
\bibfield{author}{\bibinfo{person}{S. van~den Elzen}, \bibinfo{person}{D.
  {Holten}}, \bibinfo{person}{J. {Blaas}}, {and} \bibinfo{person}{J.~J. {van
  Wijk}}.} \bibinfo{year}{2016}\natexlab{}.
\newblock \showarticletitle{Reducing Snapshots to Points: A Visual Analytics
  Approach to Dynamic Network Exploration}.
\newblock \bibinfo{journal}{\emph{IEEE Transactions on Visualization and
  Computer Graphics}} \bibinfo{volume}{22}, \bibinfo{number}{1}
  (\bibinfo{year}{2016}), \bibinfo{pages}{1--10}.
\newblock


\bibitem[\protect\citeauthoryear{Wegenkittl, L{\"o}ffelmann, and
  Gr{\"o}ller}{Wegenkittl et~al\mbox{.}}{1997}]%
        {wegenkittl1997visualizing}
\bibfield{author}{\bibinfo{person}{Rainer Wegenkittl}, \bibinfo{person}{Helwig
  L{\"o}ffelmann}, {and} \bibinfo{person}{Eduard Gr{\"o}ller}.}
  \bibinfo{year}{1997}\natexlab{}.
\newblock \showarticletitle{Visualizing the behaviour of higher dimensional
  dynamical systems}. In \bibinfo{booktitle}{\emph{Proceedings of the IEEE
  Conference on Visualization}}. \bibinfo{pages}{119--125}.
\newblock


\bibitem[\protect\citeauthoryear{Wolfe}{Wolfe}{1969}]%
        {wolfe}
\bibfield{author}{\bibinfo{person}{Philip. Wolfe}.}
  \bibinfo{year}{1969}\natexlab{}.
\newblock \showarticletitle{Convergence Conditions for Ascent Methods}.
\newblock \bibinfo{journal}{\emph{SIAM Rev.}} \bibinfo{volume}{11},
  \bibinfo{number}{2} (\bibinfo{year}{1969}), \bibinfo{pages}{226--235}.
\newblock


\bibitem[\protect\citeauthoryear{Wong and Bergeron}{Wong and Bergeron}{1994}]%
        {wong199430}
\bibfield{author}{\bibinfo{person}{Pak~Chung Wong} {and}
  \bibinfo{person}{R~Daniel Bergeron}.} \bibinfo{year}{1994}\natexlab{}.
\newblock \showarticletitle{30 years of multidimensional multivariate
  visualization.}
\newblock \bibinfo{journal}{\emph{Scientific Visualization}}
  \bibinfo{volume}{2} (\bibinfo{year}{1994}), \bibinfo{pages}{3--33}.
\newblock


\end{thebibliography}

\typeout{get arXiv to do 4 passes: Label(s) may have changed. Rerun}
\end{document}